\documentclass{article}
\usepackage[margin=1in]{geometry}
\usepackage{graphicx}
\usepackage{amsfonts}
\usepackage{subfigure}
\usepackage{booktabs} 
\usepackage{amsmath}
\usepackage{mathtools}
\usepackage{scalerel,stackengine}
\stackMath
\newcommand\reallywidehat[1]{%
\savestack{\tmpbox}{\stretchto{%
  \scaleto{%
    \scalerel*[\widthof{\ensuremath{#1}}]{\kern-.6pt\bigwedge\kern-.6pt}%
    {\rule[-\textheight/2]{1ex}{\textheight}}%WIDTH-LIMITED BIG WEDGE
  }{\textheight}% 
}{0.5ex}}%
\stackon[1pt]{#1}{\tmpbox}%
}
\usepackage{authblk}
\usepackage{grffile}
\usepackage{wrapfig,epsfig}
\usepackage{epstopdf}
\usepackage{url}
\usepackage{algpseudocode}
\usepackage[T1]{fontenc}
\usepackage{bbm}
\usepackage{comment}
\usepackage{dsfont}
\newtheorem{theorem}{Theorem}[section]

\newtheorem{definition}[theorem]{Definition}

\newtheorem{remark}[theorem]{Remark}

\newcommand{\R}{\mathbb{R}}
\newcommand{\model}{PairConnect}

\newcommand{\memmap}{\mathsf{MemMap}}
\newcommand{\onehot}{\mathsf{OneHot}}
\newcommand{\softmax}{\mathsf{softmax}}
\newcommand{\gelu}{\mathsf{GELU}}

\newcommand{\mlp}{\mathsf{MLP}}

\begin{document}
\onecolumn
\title{\bf \model: A Compute-Efficient MLP Alternative to Attention}
\date{}
\author[1]{\sl Zhaozhuo Xu}
\author[1]{\sl Minghao Yan}
\author[1]{\sl Junyan Zhang}
\author[1]{\sl Anshumali Shrivastava}
\affil[1]{Department of Computer Science, Rice University Houston, TX}
\affil[ ]{\textit {zx22, my29, jz77, anshumali@rice.edu}}

\maketitle
\vskip 0.3in
\thispagestyle{empty}

\begin{abstract} \noindent
Transformer models have demonstrated superior performance in natural language processing. The dot product self-attention in Transformer allows us to model interactions between words. However, this modeling comes with significant computational overhead. In this work, we revisit the memory-compute trade-off associated with Transformer, particularly multi-head attention, and show a memory-heavy but significantly more compute-efficient alternative to Transformer. Our proposal, denoted as {\model}, a multilayer perceptron (MLP), models the pairwise interaction between words by explicit pairwise word embeddings. As a result, {\model}  substitutes self dot product with a simple embedding lookup. We show mathematically that despite being an MLP, our compute-efficient {\model}  is strictly more expressive than Transformer. Our experiment on language modeling tasks suggests that {\model}  could achieve comparable results with Transformer while reducing the computational cost associated with inference significantly. 
\end{abstract}

\section{Introduction}

Transformer~\cite{vsp+17} has become a model of significant interest in both the research community and industry. The last few years have seen a remarkable growth of using Transformer for natural language processing~\cite{vsp+17,dclt19,ydy+19,rwc+19,bmr+20}, computer vision~\cite{dbk+20}, and recommendation systems~\cite{slw+19}. The core component of Transformer is the dot product attention mechanism.

\paragraph{The Expressive Power of Attention:}  Attention enables modeling interactions between words (or tokens) as embeddings. For illustration, consider the two phrases \emph{Apple Laptop} and \emph{Apple Fruit} where the word \emph{Apple} has totally different meanings. However, standard embedding models ignore this and assume that the representation of the word \emph{Apple} is fixed and the final representation of the phrase \emph{Apple Laptop} will be the summation (or some other pooling operation) of the embeddings of  \emph{Apple} and \emph{Laptop}. Attention goes beyond this restriction by providing a mechanism in which the embedding of the word \emph{Apple} changes based on the other words present in the sentence. When it co-occurs with \emph{Fruit}, the embedding of \emph{Apple} is very different from when it co-occurs with \emph{Laptop}. Attention, by its nature, can be recursive and hence allows for multiple layers and multiple heads, providing deeper representation. Effectively, attention, with its composability into multiple layers with numerous heads, has revolutionized natural language processing models. 

\paragraph{Attention is Expensive in Both Computation and Energy:} Formally, given a sentence, the dot product attention first computes the correlation between word embeddings. Then, the word embedding is transformed into a weighted sum of all words in the sentence with the weight represented by the correlation. 
The dot product of all-vs-all correlation is a demanding operation. The computational cost is further amplified by multiple layers and heads, where the same operation is performed repeatedly. It is well known~\cite{vsp+17,dclt19,ydy+19,rwc+19} to the community that attention is a computationally expensive, and hence energy demanding, operation. 

\paragraph{Slow Latency During Inference:} Of particular concern is the inference latency associated with Transformer. Recommendations systems are finding more and more use of large Transformer to encode queries and products~\cite{slw+19}. Thus, inference involves processing the query with a feedforward pass, including several all-vs-all dot products, to compute attention. Unfortunately, these computations do not meet the latency constraints of recommendation systems, which are generally in tens of milliseconds~\cite{slw+19,czlh19}. Furthermore, most recommendation engines are still CPU-based during inference due to the constraints in the production environment.

\paragraph{Our Key Idea: Trading Computation for Memory while Computing Embeddings:} 

To address Transformer's computation and energy burden, we use one of the most fundamental trade-offs in computer science, computation, and memory. This trade-off is already exploited in DLRM ~\cite{nms+19}. Consider vanilla embedding models where we learn embeddings from one-hot encoding~\cite{rbg+18} of a given word $w$. We can start with a one-hot encoding and pass it to a neural network to generate an embedding for $w$. Here we pay the computational cost of a feedforward network to obtain an embedding of the given word. As an alternative, which is shown in DLRM, we can store the embedding vector into embedding tables with the key $w$. The vector itself is learned. However, given $w$, retrieving its embedding is a mere memory lookup.

\paragraph{Our Contributions: {\model} }

In this paper, we propose an compute-efficient and more expressive multilayer perceptron (MLP) alternative to attention in Transformers. Our model, denoted as {\model} , uses the same observation and shows that computations in the dot product attention, with all-vs-all correlations, can be replaced by a simple pairwise embedding table lookup (hence the name {\model} ). Though we pay the price of increased memory, this is not a concern for inference on CPUs since they can afford much larger main memory. With pairwise embedding view, which we show is mathematically more expressive than dot product attention, the final network reduces to plain fully connected architecture, which is well understood by a broad community.    

We summarize our contributions as below:
\begin{enumerate}
    \item We propose a pairwise embedding table lookup method to replace the dot product attention in Transformer while maintaining comparable accuracy. 
    \item We show that our pairwise embedding table lookup method represents a more general way of modeling word interactions. 
    \item We design a neural network architecture that ingests the pairwise word embedding for language modeling. 
    \item Our pairwise modeling method is memory-heavy but significantly more compute-efficient, a huge advantage for inference in product-oriented deployment.
\end{enumerate}

\section{Related Work}
The goal of language modeling is to learn effective word representations that benefit various natural language processing tasks such as sentiment analysis~\cite{f13}, named entity recognition~\cite{lbs+16}, and machine translation~\cite{wsc16}. In previous language modeling settings, the models are required to predict the center word given its context words~\cite{bdvj03,msc+13,psm14}. Currently, a widely used paradigm is to predict the next word given previous words. For this task, neural network architectures with Long Short-Term Memory (LSTM) ~\cite{gju16}, Gated Recurrent Unit (GRU)~\cite{as17} or convolutional ~\cite{dfag17} layers are proposed with promising results.

In the last few years, we have witnessed the remarkable growth of using Transformer to significantly improve the state-of-the-art in natural language processing~\cite{vsp+17,dclt19,ydy+19,rwc+19,bmr+20}. The major components of Transformer could be summarized as three steps: word embedding, word interaction modeling, and task-driven prediction. The word embedding phase first tokenizes the sentence and represents each token with a latent vector. Next, positional encoding~\cite{vsp+17} is performed over the sequences of tokens by adding sine/cosine functions over the value of latent vectors. After the word embedding phase, Transformer models the iteration of words in a sentence by the dot product attention model, which is defined as:
\begin{definition}[Attention~\cite{vsp+17}]
Given input sentence embedding $X\in \R^{n\times d_s}$, we apply 3 separate feedforward layers to generate embeddings $Q\in\R^{n\times d}$, $K\in\R^{n\times d}$, and $V\in\R^{n\times d}$. The attention is represented as 
\begin{align*}
    head=\softmax\left(\frac{QK^T}{\sqrt{d}}\right)\cdot V
\end{align*}
where $\softmax$ denotes the softmax function.
\end{definition}

Finally, the token with transformed embeddings is fed to MLP layers to predict the next word~\cite{vsp+17}, the masked words~\cite{dclt19,ydy+19,rwc+19,bmr+20}, and/or the translated words~\cite{vsp+17,dclt19}.

The development of Transformer also brings up new tasks for language modeling. In~\cite{dclt19}, a masked language modeling task is introduced. In this task, the neural network is tasked to predict the masked words given the context. Currently, masked language modeling has become an important natural language processing task. 

Although attention-based Transformer models achieve the state-of-the-art performance in language modeling, the training of Transformer consumes massive amount of time and energy, which prevents the research community from engaging in this line of research. Therefore, a series of efficient Transformers is proposed to accelerate the computation and to reduce the training overhead~\cite{kkl19,clv20,tbm+20,wlk+20,clp+20,laeo21}. However, all current approaches follow the paradigm of modifying the attention formulation. There are two fundamental questions yet to be answered: (1) Can we replace the weighted sum styled attention block with direct pairwise word level modeling? (2) Can we replace the expensive matrix multiplications with efficient embedding lookups?

\section{Approach}
In this section, we present the architecture of {\model} . We start with showing a direct connection between embedding table lookups used in Facebook's DLRM~\cite{nms+19} and fully connected networks over one-hot encoding. This connection highlights that many seemingly different architectures are essentially different variants of memory-compute trade-off. Moreover, it brings insights to language modeling.  We then present our central idea, the pairwise word embedding. Next, we show how to achieve multi-head and multi-layer pairwise word modeling in {\model} . Finally, we provide the architecture of {\model} .
\begin{figure}
\centering
\includegraphics[width=1\linewidth]{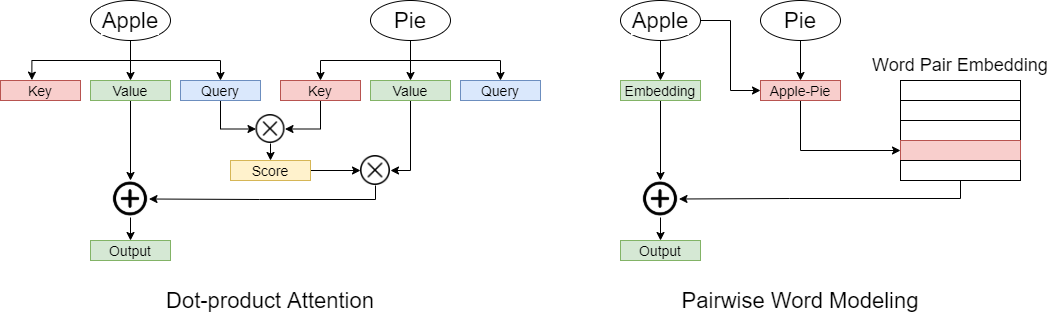}
\caption{Comparison of pairwise word modelling and attention}
\label{fig:compare}
\end{figure}
\subsection{Starting Observation: Embedding Lookup in DLRM~\cite{nms+19} is Equivalent to Fully Connected Model with One-hot Encoding}

DLRM~\cite{nms+19} model deploys embedding tables for categorical data, including words. Let $U$ be the size of our vocabulary $V = \{w_1, w_2, ...., w_U\}$, where each $w_i$ is a word. DLRM model maps every word $w_i$ to a corresponding embedding vector $E_i$ in the embedding table. The embedding table is simply a $(key, \ value)$ dictionary with key being the word $w_i$ and value being the corresponding embedding $E_i$. The embeddings $E_i$ are directly learned from data.

Consider a sentence of $n$ words $S = \{x_1, \ x_2, \ ..., \ x_n \}$, where $x_i \in V$.  Since the embedding $E_i$ is learned, we will treat them as a learned function of the corresponding word $x_i$, denoted by $f_{\memmap}(x_i)$. We call it memory map as computing $f_{\memmap}(x_i)$ is simply a dictionary lookup which does not require any arithmetic computations. The embedding of a sentence is the summation (or some other pooling operation) given by $\sum_{i=1}^n f_{\memmap}(x_i)$. Due to deep neural network's representative power, the embedding $f_{\memmap}(.)$ can be any complex function we wish.  

Now consider the standard fully connected layer on the one-hot encoding of sentence $S$. Such a fully connected network is also known as DSSM (by Amazon)~\cite{hhg+13}. The one-hot encoding of $S$, given the vocabulary $V$, can be written as a binary vector. Let us denote it by $\onehot(S)$. $\onehot(S)$ is of dimension $U$ (the size of vocabulary). In $\onehot(S)$, the components are 1 for present words and 0 for absent words. A fully connected layer multiplies a weight vector $W \in R^{k \times U}$ with $\onehot(S) \in R^{U \times 1}$, where $k$ is the embedding dimension. Then, the layer performs a non-linear activation function on the result. The output of the layer can be written as $\sum_{i=1}^n W_{x_i}$, where $W_{x_i}$ represents the embedding of $x_i$ in $W$. Here the $x_i^{th}$ column corresponds to the word $x_i$. Clearly, this is identical to DLRM if we set $f_{\memmap}(x_i) = W_{x_i}$. There is not much difference between the form of a fully connected layer over one-hot encoding and embedding tables. The non-linearly after matrix multiplication in a fully connected layer can be absorbed in the $f_{\memmap}$ function in the embedding table lookup method. However, the difference shows up when we include hardware and computations in the loop.

\paragraph{Compute or Memory Lookup: Which is better?} Since one-hot is a sparse binary vector, performing dictionary (or memory) lookup and addition is faster than matrix multiplication on CPUs. However, in a modern High Performance Computing (HPC) environment, the choice depends. For example, on GPUs, sparse memory lookups are costly, while matrix multiplications may be cheaper in a batch. It is not a big surprise that Facebook trains DLRM models on CPUs~\cite{nms+19} and Amazon trains DSSM on GPUs~\cite{hhg+13}. 

To summarize, if we consider deep learning a function transformation, we can either compute the transformation every time we need it, or memory map it and learn the map's values. The choice depends on the platform. However, in the case of attention, the computations are significantly heavier. 

\subsection{Pairwise Word Embedding Lookup as An Alternative to Attention}
We start by taking the functional view of attention embedding, which will help us define the compute-memory trade-off more efficiently. Consider a sentence $S$ consisting of $n$ words, $S = \{x_1, \ x_2, \ ..., \ x_n \}$, where $x_i \in \R^d$ is a unigram word embedding. We denote the $K$, $Q$ and $V$ embedding for a given word $w_i$ as functions, or $f_K(x_i)$,  $f_Q(x_i)$, and $f_V(x_i)$, respectively. With this notation, we can define the attention embedding $AE$ of word $x_i$ as  $$AE(x_i) = \sum_{j=1}^n \softmax(\frac{f_Q(x_i)^\top f_K(x_j)}{\sqrt{d}}) \times f_V(x_j).$$

We can effectively absorb all functions in $AE(x_i)$'s expression under a binary function \begin{equation}
\label{eq:Form}
F_{\memmap}(x,y) = \softmax(\frac{f_Q(x)^\top f_K(y)}{\sqrt{d}}) \times f_V(y)\end{equation} 
and rewrite $AE(x_i)$ as 
$AE(x_i) = \sum_{j=1}^n F_{\memmap}(x_i,x_j)$.
Here $F_{\memmap}(x_i,x_j)$ represents the joint embedding of the ordered pair $x_i, \ x_j$. It should be noted that the ordering matters as $F_{\memmap}(x_i,x_j) \ne F_{\memmap}(x_j,x_i)$.
Moreover, similar to DLRM~\cite{nms+19}, we will store these embeddings in embedding tables. Therefore, computing $F_{\memmap}$ only requires a simple dictionary (or memory) lookup. Since we are learning the output of the embedding $F_{\memmap}$ directly, we don't have to perform the dot product computations.

\paragraph{Expressive Power of Pairwise Word Embeddings} Since we directly learn the output of $F_{\memmap}(x_i,x_j)$, we can potentially learn any binary function. Clearly, attention requires a specific decomposition for this binary function, given by Equation~\ref{eq:Form}, and hence is more restrictive than our method, which does not require a decomposition.

\begin{remark}
Assuming that the learning operation can generate any functional mapping, Pairwise Word Embeddings are strictly more Expressive than Attention. We simplified the mathematics by assuming a function $f$ for embedding of word $w_i$ as $f(w_i)$. However, when the embeddings are learned directly, we can essentially have a different function for every word (or word pair). The arguments presented do not change even if we remove this simplification. 
\end{remark}

% {\bf Note:} We simplified the mathematics by assuming a function $f$ for embedding of word $w_i$ as $f(w_i)$. However, when the embeddings are learned directly, we can essentially have a different function for every word (or word pair). The arguments presented do not change even if we remove this simplification. 

\subsection{Details of the Memory Map $F_{\memmap}$ and other Transformation}

In this section, we show how to learn the proposed memory map function $F_{\memmap}$. The objective of $F_{\memmap}$ is to directly represent each word pair via a latent vector and avoid attention calculation explicitly. For instance, given a sentence "apple pie", we would directly pick the embedding of word pair "apple-pie" (as shown in Figure~\ref{fig:compare}). Note that the general paradigm for word embedding table is close to~\cite{cw08}. 

% \Zhaozhuo{Draw a picture about pairwise modeling}

We process the embedding from the table with  basic $\mlp$ blocks, which is an essential component for learning. We use 2-Layer $\mlp$ to project the pairwise embedding for target oriented modeling, defined as below:

\begin{definition}[2-Layer $\mlp$ ]\label{def:2mlp}
Let $x\in\R^{m\times d }$ denote an input matrix for the layer, where $m$ is the number of embeddings and $d$ is the dimension of the embedding. Let $W_0\in R^{d\times d_1}$ and $W_1\in R^{d_1\times d_2}$ denote the weights for two linear layers, respectively. Let $\gelu$ denote the Gaussian Error Linear Units (GELU) activation function~\cite{hg16}. Let $\sigma$ denote the dropout function. We denote a 2-layer $\mlp$ as:
\begin{align*}
    f(x)=\sigma(\sigma( \gelu(xW_0))W_1)
\end{align*}
\end{definition}

As shown from Definition~\ref{def:2mlp}, the $\mlp$ uses Gaussian Error Linear its (GELUs) activation function~\cite{hg16} after two linear projections. Next, we show that the function $F_{\memmap}$ is represented as:
$$F_{\memmap}(x_i,x_j)=f(W_{ij})$$
where $W$ is the embedding table that stores the embedding for every pair of words in the vocabulary.

Finally, the embedding in $\R^d$ for all word pairs in this sequence can be summarized as a function $g$ as:
\begin{align}\label{eq:pair_model}
    g(x_i)=\sum_{j=1}^{m}F_{\memmap}(x_i,x_j)
\end{align}
% \begin{definition}[Pairwise Modeling Layer]\label{def:pair_model}
% Let $m$ denote the total sequence length. Let $f(x):\R^{m\times d}\rightarrow\R^{m\times d_2}$ denote the 2 layer MLP (see Definition~\ref{def:2mlp}).
% Let $X_i=\{x_{i1},x_{i2},\cdots, x_{im}\}$ denote the set of all word pairs for the $i$th word in a sequence, where each element is a tuple of word indices. Given the compressed pairwise word embedding (see Definition~\ref{def:cpwel}) with embeddings $W\in \R^{d\times k}$ and universal hash function $h: \mathbb{N} \rightarrow [k]$, the pairwise modeling layer for input sequence is defined as 
% \begin{align*}
%     g(x)=f_1(f(W(X))) 
% \end{align*}
% where $W(X)\in \R^{m\times d}$ is the embedding for word pairs extracted from the pairwise embedding table, and $f_1:\R^{m\times d}\rightarrow \R^{d}$ is the function that takes the sum of a matrix over columns.
% \end{definition}

The pairwise modeling layer generates a vector in $\R^d$ by summing all the word pair embeddings after the 2-layer $\mlp$. In this way, we directly model the iteration between words in a sentence. In the next section, we build multi-layer and multi-head extensions to use this pairwise word representation, which will provide us with the depth and the breadth.

\subsection{Multi-head and Multi-Layer Pairwise Modeling}

% After we develop the pairwise modeling layer (Definition~\ref{def:pair_model}), we are able to propose a novel multi-head and multi-layer language modeling architecture that goes beyond dot-product attention
A nice property of attention layer is that it consumes an embedding and generates another one which can be applied recursively (multi-layer) to make the model deeper and in parallel to make the model wider (multi-head). This depth and breadth provides the opportunity for refinement which is crucial for improving the accuracy (Also see our experiments Section~\ref{sec:multi-head} where we compare the effect of multi-head and multi-layer attention side by side with attention). 

Since pairwise embedding model also has the same property, the extension to multi-head and multi-layer is straightforward.  

For multi-head we initialize the function $g$ in Eq~\eqref{eq:pair_model} $l$ times with different parameters. The multi-head pairwise modeling layer is denoted as function $\hat{g}$ with form
\begin{align}\label{eq:combine}
    \hat{g}(x_i)=[g_1(x_i)^\top, g_2(x_i)^\top,\cdots,g_l(X_{x_i})^\top]^\top.
\end{align}
This multi-head design is an analogy to the multi-head attention in Transformer~\cite{vsp+17}. \cite{vsp+17} performs several dot product attention blocks and concatenates their output vectors. Instead, we only need multiple embedding tables and lookups. 

The  multi-layer version of {\model}  is pretty straightforward. As shown in Figure~\ref{fig:multi-head}, the output of one pairwise word modeling block has the same size as the input. In other words, the output of pairwise modeling block is a new unigram embedding for the sentence. The next layer then uses independent pairwise embedding tables to refine this embedding again before passing forward.

\subsection{{\model}  Architecture}

\begin{wrapfigure}{r}{0.5\textwidth}
\centering
\includegraphics[width=.8\linewidth]{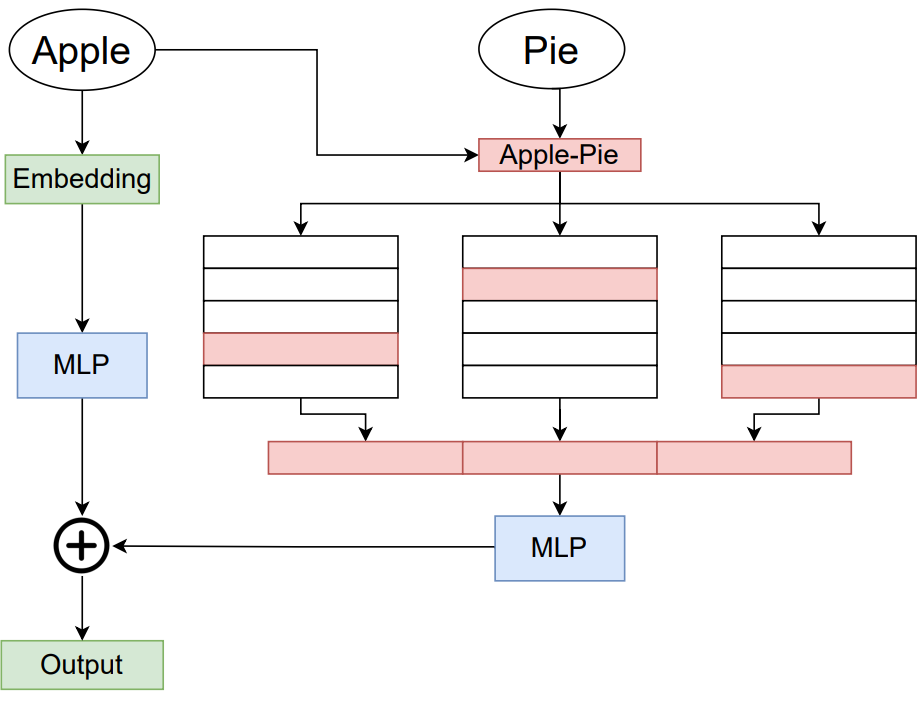}
\caption{Illustration of multi-head {\model}  layer. Pairwise word embeddings from multiple tables are extracted and concatenated. Then, PariNet passes it through and MLP and sum it with original word embedding. The output servers as input embedding for next layer with independent embedding table for pair.}
\label{fig:multi-head}
\vspace{-4mm}
\end{wrapfigure}

In this section, we present the full architecture of {\model} . The {\model}  architecture contains several layers. As shown in Figure~\ref{fig:multi-head}, in each layer, given the unigram embedding of a sentence, we first lookup all pairwise embeddings from multiple embedding tables. Each embedding table represents one head. Then, for each unigram word embedding $x\in \R^d$, one embedding table prepares pairwise embeddings $x_p\in \R^{(m-1)d}$ for this word, where $m$ is the sequence length. The pairwise word embedding is obtained following Eq.~\eqref{eq:pair_model}. We could also repeat the pairwise modeling and then concatenate the output vectors from each table following Eq.~\eqref{eq:combine}.  Next, we project $x$ into $x'\in \R^d$ and $x_p$ to $x'_p\in \R^d$ via separate MLPs. After that, we take the summation of $x'_p$ and $x'$ and feed to another MLP layer. In this paradigm, we learn the unigram word embedding with same dimension as input through this one-head {\model}  layer.

\subsection{Feature Hashing to Handle Quadratic Memory Blow-up}
% \begin{figure}

Given the language dataset with vocabulary size $U$, a naive way of modeling the pairwise relationship between words is to build a weight matrix $W\in \R^{d\times U^2}$ with each column representing the pairwise word embedding. However, this type of modeling is infeasible in current computational machines as the $U^2$-column matrix would exhaust the memory. To tackle this issue, we propose a compression method based on universal hashing~\cite{cw79}, also known as feature hashing or simply the hashing trick~\cite{wdl+09}.

In {\model} , we set the embedding table size to be $K$. Given each pair of words, we randomly hash the string into a positive integer less than $K$ through MurmurHash~\cite{m08}. Then, we look up the embedding according to the hash value. This way, we map each word pair to a latent vector, where the probability that two-word pairs have the same embedding is equal to $1/K$. Note that we reduce this collision probability by multi-head pairwise modeling. In each head, we choose independent randomized hash functions. Therefore, the collision probability that two-word pairs have the same embedding after $l$ heads is at most $1/K^{l}$. With this tiny collision probability, the multi-head pairwise modeling layer can represent complicated word relationships in a sentence.

Theoretically, the hashing trick is well known~\cite{wdl+09} to not affect the accuracy of learned models. It provides a sample mapping from tokens to embeddings.  We observe in experiments (Section~\ref{sec:abla}) that with hashing trick, we can work with significantly smaller embedding table size without any loss in accuracy.

\subsection{Computational Efficiency}

% \vspace{-5mm}
In this section, we illustrate that {\model}  is an efficient neural network architecture for language modeling. One major system level feature of {\model}  is that it transforms matrix multiplication into an embedding lookup. It is well known that the major computational bottleneck of Transformer is its expensive self matrix multiplication. Given the word embedding $X\in \R^{m\times d}$, the dot product attention requires $XX^\top$, which takes $O(m^2d)$. In {\model} , we go beyond this step by directly looking up the pairwise embeddings from the embedding tables, which costs $O(d)$ per embedding lookup.

There are several major advantages of the embedding lookup based pairwise word modeling over dot product attention. First, it reduces the massive matrix multiplications during the attention blocks in Transformer. Second, the embedding lookup could be further accelerated on CPUs, which benefits efficient inference of language models on CPUs. Note that {\model}  requires more memory during training than Transformer as we directly model the word pairs with embedding tables. We will demonstrate in more details in the experiment section that the hashing algorithm can compress the tables to suitable sizes so that we would not exhaust the memory.

\section{Experiment}
In this section, we present the experiment results for {\model}. Specifically, we would like to answer the following questions.
\begin{itemize}
    \item Is the proposed {\model} comparable with Transformer in accuracy for language modeling? More specifically, we also compare {\model} with attention by varying the number of heads and layers to confirm that the power of {\model} is similar to that of attention mechanism.
    \item What are the major hyper-parameters in {\model}? Is {\model}'s performance sensitive to those hyper-parameters?
    \item How efficient is {\model} when comparing to Transformer during inference?
\end{itemize}

In the following sections, we first present the settings and results for masked language modeling. Then, we show an ablation study on the parameters of {\model}. Finally, we present the running time of {\model} in the inference phase with comparison to Transformer.
\subsection{Masked Language Modeling}
In this section, we present our main results of {\model} on masked language modeling (MLM). MLM is a standard task for popular Transformer-based language models such as BERT~\cite{dclt19}, XL-Net~\cite{ydy+19}, and the GPT series~\cite{rwc+19,bmr+20}. In MLM, given a sentence represented as a sequence of words, we randomly mask a subset of words. Then, the goal of MLM is to predict the masked words given the processed sentence. 

\paragraph{Dataset} 
In this work, we present the masked language modeling results on two most popular datasets: Penn Treebank (PTB)~\cite{mz12} and Wiki-Text-2~\cite{scjr16}. For each dataset, we set the padding token id to be $0$ and the masked token id to be 1. Then, we perform tokenization on words starting from 2. Next, we follow the pattern in \cite{dclt19} and randomly mask $15\%$ of words in each sequence with the id 1. Note that only $90\%$ of sequences would be randomly masked. We follow the standard train/test split in \cite{mz12} and \cite{scjr16}. 

\paragraph{Setting}
We implement {\model} in PyTorch~\cite{pgm+19}. We compare {\model} with Transformer implemented in PyTorch on the two datasets. The experiment is conducted on a server with 8 Nvidia Tesla V100 GPU and two 20-core/40-thread processors (Intel Xeon(R) E5-2698 v4 2.20GHz). 

\paragraph{Evaluation Metrics}
We use the standard cross-entropy loss to train {\model} and Transformer. When training converges, we evaluate the training and testing losses as an accuracy metric. 

\paragraph{Parameters}
For both {\model} and Transformer, we set number of layers to 6, number of heads to 4, word embedding dimension to 256, and hidden dimension to 256. We train both models using Adam optimizer~\cite{kb14} and set the learning rate to be 1e-5. 

\subsubsection{Head-to-Head Evaluation of Transformer and {\model} with Varying Number of Heads and Layers} \label{sec:multi-head}

In this section, we present the accuracy evaluation of {\model} and its comparison with Transformer. As shown in Table~\ref{tab:main}, {\model} achieves the same evaluation loss as Transformer in MLM while having a slightly lower training loss in the PTB dataset. In the Wiki-Text-2 dataset, {\model} performs close to Transformer and has a slightly higher testing loss. These results answer the first question partially. In single-layer and single-head setting, {\model} can achieve an accuracy comparable with Transformer without using the dot product self-attention layer. 

We also extend both {\model} and Transformer to multi-layer and multi-head settings and present the performance of {\model} and Transformer as we increase number of heads and number of layers. We increase the two hyper-parameters till we exhaust the memory of one Nvidia V100 GPU. We observe that: (1) Both {\model} and Transformer's performance improve as we increase number of heads and number of layers. (2) {\model} achieves comparable results with Transformer during each set of parameters. These results indicate that {\model} could achieve similar performance with Transformer when scaling up to multi-layer and multi-head settings. As the 6-layer and 4-head version of Transformer achieves the state-of-the-art MLM performance on both datasets, we validate the effectiveness of {\model}.

\begin{table}[h]
\begin{center}
\setlength{\abovecaptionskip}{3pt}
\setlength{\belowcaptionskip}{2pt}
 \caption{Results for Mask Language Modeling}
 \begin{tabular}{ p{2cm}|p{2cm}|p{1.5cm}|p{1.5cm}|p{2cm}|p{2cm} }
 \hline
 Dataset &  Models & {$\mathbf{\#}$ Layers} & {$\mathbf{\#}$ Heads}& Training Loss & Test Loss  \\
 \hline\hline
% Simulate & 1600 & 100 & 201$\times$301  \\
PTB & {\model} & 1 & 1 &  1.39 &  1.49  \\
PTB & {\model} & 1 & 2 &  1.39 &  1.47  \\
PTB & {\model} & 2 & 2 &  1.22 &  1.39  \\
PTB & {\model} & 2 & 4 &  1.16 &  1.25  \\
PTB & {\model} & 4 & 4 &  1.05 &  1.13  \\
PTB & {\model} & 6 & 4 &  0.98 &  1.09  \\\hline
PTB & Transformers & 1 & 1 &  1.4 &  1.49  \\
PTB & Transformers & 1 & 2 &  1.36 &  1.48  \\
PTB & Transformers & 2 & 2 &  1.19 &  1.37  \\
PTB & Transformers & 2 & 4 &  1.17 &  1.29  \\
PTB & Transformers & 4 & 4 &  1.04 &  1.13  \\
PTB & Transformers & 6 & 4 &0.99 & 1.09  \\ \hline
Wiki-Text-2 & {\model} & 1 & 1 &  1.66 &  1.76  \\
Wiki-Text-2 & {\model} & 1 & 2 &  1.53 &  1.69  \\
Wiki-Text-2 & {\model} & 2 & 2 &  1.29 &  1.44  \\
Wiki-Text-2 & {\model} & 2 & 4 &  1.09 &  1.19  \\
Wiki-Text-2 & {\model} & 4 & 4 &  1.06 &  1.15  \\
Wiki-Text-2 & {\model} &6 & 4 & 1.05 &  1.12 \\\hline
Wiki-Text-2 & Transformers &1 & 1 & 1.67 & 1.75  \\
Wiki-Text-2 & Transformers &1 & 2 & 1.56 & 1.71  \\
Wiki-Text-2 & Transformers &2 & 2 & 1.28 & 1.42  \\
Wiki-Text-2 & Transformers &2 & 4 & 1.11 & 1.25  \\
Wiki-Text-2 & Transformers &4 & 4 & 1.04 & 1.17  \\
Wiki-Text-2 & Transformers &6 & 4 & 1.02 & 1.11  \\
 \hline
 
\end{tabular}\label{tab:main}
\end{center}
\end{table}

\subsubsection{Ablation Study for the Hashing Trick and Table Size}\label{sec:abla}

One unique hyper-parameter of {\model} is the hash size for pairwise word embeddings. In this section, we vary the hash size $K$ of the {\model} with 6 layers and 4 heads from 100 to 10000 and obtain the following results. These results answer the second question. {\model} achieves stable accuracy with different hash sizes. 

\begin{table}[h]
\begin{center}
\setlength{\abovecaptionskip}{3pt}
\setlength{\belowcaptionskip}{2pt}
 \caption{Ablation Study of {\model}}
 \begin{tabular}{ p{2cm}|p{2cm}|p{2cm}|p{2cm}|p{2cm} }
 \hline
   Hash Size ($K$)& Dataset  & Test Loss & Dataset & Test Loss  \\
 \hline\hline
 100 &PTB &  1.1 &  Wiki-Text-2   &  1.19  \\
500 &PTB &  1.09 &  Wiki-Text-2  &  1.12  \\
1000 &PTB &  1.09 &  Wiki-Text-2   &  1.12  \\
5000 &PTB &  1.1 &  Wiki-Text-2   &  1.13  \\
 10000 &PTB &  1.1 &  Wiki-Text-2  &  1.2  \\
 \hline
\end{tabular}\label{tab:abla}
\end{center}
\end{table}

\subsection{Inference Efficiency on CPUs}

Due to heavy energy consumption, most of Deep Learning applications are deployed on CPUs. Therefore, even if the model is trained on GPUs, inference speed on CPU is a major evaluation metric for its efficiency. As shown in previous sections, one of the major advantages of {\model} is that it trades memory for efficiency and replaces the expensive and slow matrix multiplications with cheap and fast embedding lookups. In this section, we present the experimental results on the comparison between inference on {\model} and Transformer. 
\paragraph{Setting:}
All the experiments are conducted on a single thread of a machine equipped with two 20-core/40-thread processors (Intel Xeon(R) E5-2698 v4 2.20GHz). The machine is installed with Ubuntu 16.04.5 LTS. The full inference is implemented in PyTorch~\cite{pgm+19}.

\paragraph{Efficient Embedding Lookup:}
We implement our hash tables and hash table lookups in C++ with PyTorch~\cite{pgm+19} wrappers to fully take advantage of its speed. Since each bucket in a hash table stores an embedding, as opposed to a normal hash table where there might be multiple items stored in a single bucket, we implement each hash table as a giant contiguous array.

To retrieve an embedding from a hash table, we support both passing by reference and passing a new array where we concatenate all retrieved embeddings. Passing a series of pointers to the head of the retrieved embeddings saves the operation of copying the retrieved embeddings to a new array but requires more memory lookups, and vice versa. In practice, users can choose the strategy that gives the best latency. We use the second method in our experiments.
\begin{table}[h]
\begin{center}
\setlength{\abovecaptionskip}{3pt}
\setlength{\belowcaptionskip}{2pt}
 \caption{Results on {\model} Inference. The Speed denotes the number of samples inferred in one second. Transformer can process 45.35 samples per second.}
 \begin{tabular}{ p{2cm}|p{2cm}|p{2cm}|p{2cm}|p{2cm}|p{2cm} }
 \hline
 Hash size&  100 & 500 & 1000 & 5000& 10000  \\
 \hline\hline
 Speed & 62.51  & 58.82  & 55.55  & 52.63  & 50  \\
 \hline
\end{tabular}\label{tab:inf}
\end{center}
\end{table}
\paragraph{Results:} We perform CPU inference of {\model} and Transformer on the PTB dataset. In this experiment, we set the batch size to 1 and perform inference on a single thread. Both {\model} and Transformer are built with 6 layers and 4 heads. As shown in Table~\ref{tab:inf}, we measure the average number of samples processed per second. The Transformer model could process 45.35 samples per second on average. Our {\model} outperforms Transformer by processing more samples. When hash size is set to 1000, {\model} achieves the best test loss and is 22$\%$ faster than Transformer in CPU inference. Meanwhile, we observe that {\model}'s inference speed decreases as hash size increases. This is expected since larger hash size would mean more hash table lookups. These results answer the third question: {\model} is faster than Transformer in CPU inference.

\vspace{-2mm}
% \section{Negative Societal Impact}
% This paper proposes an efficient language modeling approach. However, training {\model} , just like other heavy ML models, has environmental impact and is likely to consume a significant amount of energy. Future work will focus on making training energy efficient, and thus, reducing its negative environmental impact.
% \vspace{-2mm}
\section{Conclusion}
In this work, we identify that though Transformer-based models have demonstrated superior performance in natural language processing, they introduce significant computational overhead in dot product self-attention. To tackle this issue, we revisit the memory-compute trade-off associated with Transformer and propose a memory-heavy but significantly more compute-efficient alternative to Transformer. Our proposal, {\model} , is a multi-layer perceptron (MLP) that models pairwise interaction between words by explicit pairwise word embeddings. As a result, {\model}  substitutes self dot product with a simple and efficient embedding lookup. We show that despite being an MLP, our compute-efficient {\model}  is strictly more expressive than Transformers. Our experiment on language modeling task suggests that {\model}  could achieve comparable results with Transformer while reducing the inference cost. 

\bibliography{ref}
\bibliographystyle{unsrt}

\end{document}